\documentclass[letterpaper, 10 pt, conference]{ieeeconf}  
\IEEEoverridecommandlockouts                              
\usepackage{graphicx}
\usepackage{subfigure}
\usepackage{amsmath}
\usepackage{amssymb}
\usepackage{array}
\usepackage{multirow}
\usepackage{multicol}
\usepackage{booktabs}
\title{\LARGE \bf
Prioritized Planning for Target-Oriented Manipulation via Hierarchical Stacking Relationship Prediction
}

\author{Zewen Wu$^{1}$, Jian Tang$^{1}$, Xingyu Chen$^{1}$, Chengzhong Ma$^{1}$, Xuguang Lan$^{1}$ and Nanning Zheng$^{1}$
\thanks{$^{1}$Z. Wu, J. Tang, X. Chen, C. Ma, X. Lan, and N. Zheng are with National Key Laboratory of Human-Machine Hybrid Augmented Intelligence, National Engineering Research Center for Visual Information and Application, Institute of Artificial Intelligence and Robotics, Xi'an Jiaotong University, No.28 Xianning Road, Xi’an, Shaanxi, China 
{\tt\small wuzewenchn@foxmail.com, xglan@mail.xjtu.edu.cn}}
}

\begin{document}

\maketitle
\thispagestyle{empty}
\pagestyle{empty}

\begin{abstract}

In scenarios involving the grasping of multiple targets, the learning of stacking relationships between objects is fundamental for robots to execute safely and efficiently. However, current methods lack subdivision for the hierarchy of stacking relationship types. In scenes where objects are mostly stacked in an orderly manner, they are incapable of performing human-like and high-efficient grasping decisions. This paper proposes a perception-planning method to distinguish different stacking types between objects and generate prioritized manipulation order decisions based on given target designations. We utilize a Hierarchical Stacking Relationship Network (HSRN) to discriminate the hierarchy of stacking and generate a refined Stacking Relationship Tree (SRT) for relationship description. Considering that objects with high stacking stability can be grasped together if necessary, we introduce an elaborate decision-making planner based on the Partially Observable Markov Decision Process (POMDP), which leverages observations and generates the least grasp-consuming decision chain with robustness and is suitable for simultaneously specifying multiple targets. To verify our work, we set the scene to the dining table and augment the REGRAD dataset with a set of common tableware models for network training. Experiments show that our method effectively generates grasping decisions that conform to human requirements, and improves the implementation efficiency compared with existing methods on the basis of guaranteeing the success rate.

\end{abstract}

\section{INTRODUCTION}\label{intro}

Robot grasping in stacking scenarios has always been a challenging problem. When performing manipulation tasks, robots are required to execute efficiently and ensure the environments are controllable and accident-proof. Therefore, it is indispensable to learn the stacking relationships between objects \cite{liu2012fast} \cite{murali20206}. Considering a kind of multi-object stacking scenario, where objects are not only in a stacked state but the object below provides full support for the object above, and the latter reaches a stable equilibrium state. This situation widely exists in fields of dish serving, kitchen tidying, logistics transporting, and so on. It is a special case of clutter \cite{fischinger2013learning}, while its unique attribute contributes to the high speed and efficiency of tasks on the premise that the robot fully understands the environment.

\begin{figure}[t]
  \centering
  \subfigure[]{
  \includegraphics[width=8.5cm]{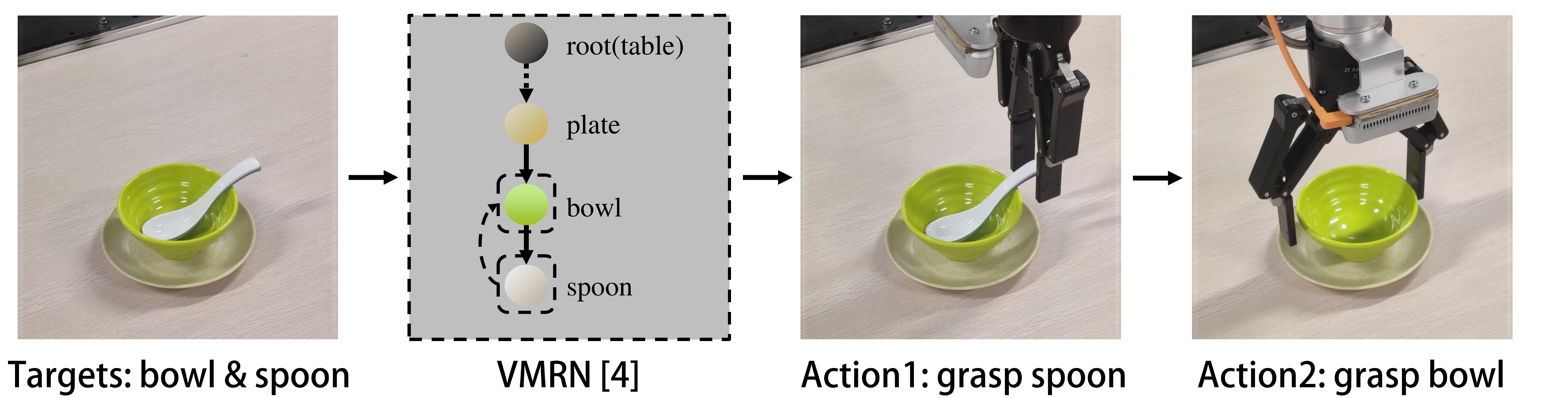}}
  \quad
  \subfigure[]{
  \includegraphics[width=8.5cm]{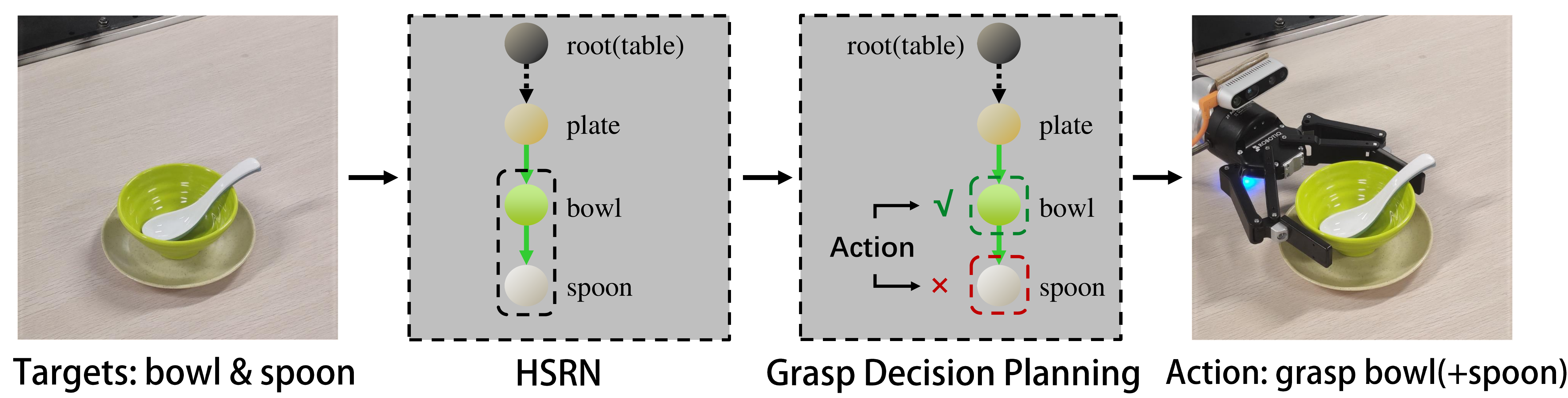}}
  \caption{Grasping in the orderly-stacked scene. The staking relationships are presented by a tree structure. When the spoon is stably supported by the bowl and both are the targets, grasping the bowl directly will be more in line with human common sense and improve execution efficiency. ($\textit{a}$) Existing method \cite{VMRN} generally perform them separately. ($\textit{b}$) Our perception-planning method optimizes the grasping decision, only needs to grasp the bowl to complete the task.}
  \label{fig1}
\end{figure}

Existing algorithms in learning stacking relationships for manipulation focus on distinguishing simple relationship types. Zhang et al. \cite{VMRN} establish Visual Manipulation Relationship Network (VMRN) to predict the manipulation relationship of each object-pair in the scene, combine them as a tree structure for presenting results so that a grasping order can be determined when any target is specified, as shown in Fig. \ref{fig1}($\textit{a}$). The relevant works \cite{ggnn-vmrn} \cite{tchuiev2022duqim} optimize the manifestation and performance of relationship detection, but lack further subdivision for the hierarchy of stacking relationship types. In scenarios where objects are stably stacked, it is insufficient for the robot to perform high-efficient grasping.

Inspired by human common sense, our work follows a principle: when we want to grasp a pile of objects that are stably stacked together, we can directly grasp the bottom object, as illustrated in Fig. \ref{fig1}($\textit{b}$). For example, when assigning a robot to grasp and deliver a pile of dishes and bowls, it is irrational to perform them separately, but grasp once and for all. This condition will become more intricate when one of the dishes is not required. Another case is when there are multiple sundries stably stacked above the target object, removing the sundries all at once will greatly save time and the number of manipulations. Therefore, we try to solve the problem that how to decide the best grasping action in the current orderly stacking scenario when targets are designated.

In this paper, we present a perception-planning architecture for generating optimal grasping decisions when specifying any number of objects in the scene as targets. We introduce a Hierarchical Stacking Relationship Network (HSRN) for scene perception, which takes dining scene RGB images as input, and outputs the hierarchical stacking relationship between each pair of objects through object detection and relationship prediction, presented by the Stacking Relationship Tree (SRT). Then a planning algorithm designed from the Partially Observable Markov Decision Process (POMDP) is implemented to comprehensively consider the inadequacy of the perceptual process and the priority of the grasping targets, obtain an underlying estimate of the real state and provide an optimal grasping decision chain. In order to train our network, we augment a virtual orderly-stacked scene dataset on the basis of REGRAD \cite{regrad}, attempting to simulate the stacking environment of dining tables. Experimental results substantiate the feasibility of our perception and planning process, which outperforms state-of-the-art baselines in target-oriented task manipulation while saving considerable time. Our main contributions are summarized as follows:

\begin{itemize} 
\item A hierarchical stacking relationship prediction network that distinguishes different degrees of stacking between objects.
\item A POMDP-based planner for planning optimal manipulation decision chain in the current scene when targets are specified.
\item The experimental results confirm that our method improves the efficiency of performing grasping tasks, and is more in line with human habits.
\end{itemize}

\section{RELATED WORK}

\subsection{Visual Stacking Relationship for Manipulation}

The visual relationship has been studied in the field of computer vision for a long time. In analyzing the relationship between the foreground in the image, the classification consideration is frequently based on the triplets of $<$subject, predicate, object$>$ as a whole \cite{farhadi2011recognition} \cite{divvala2014learning}. Other works cope with this problem by predicting subject, object, and their relationship separately \cite{lu2016visual} \cite{liang2018visual}. However, this kind of relationship analysis lacks direct guidance for robotic manipulation. 

In the scene of stacking cluttered objects, the visual relationship between objects is mainly represented as support and occlusion. Panda et al. \cite{panda2013learning} define a variety of support relationships between objects, including support from below, support from the side, and containment, constructing the inferred support sequence of objects. Zhang et al. \cite{VMRN} build a Visual Manipulation Relationship Network (VMRN) for representing the stacking relationship in the scene, and collect a dataset for relationship detection in robotic grasping. Yang et al. \cite{crf} introduce fully connected Conditional Random Fields (CRFs) on \cite{VMRN}, removing redundant relationship representations. Zuo et al. \cite{zuo_graph} and Ding et al. \cite{ggnn-vmrn} deploy graph neural networks to collect contextual information about objects and improve relationship detection performance. Tchuiev et al. \cite{tchuiev2022duqim} utilize Deformable DETR \cite{zhu2020deformable} as the backbone, representing object hierarchy in directed graph adjacency matrix form. Current approaches focus on pursuing high accuracy of detection, but for orderly-stacked scenarios, a lack of in-depth understanding of the relationship types cannot satisfy diversified task requirements. In comparison, our work unveils subtle classification and verification of the stacking relationship, which is more in line with human cognitive habits for manipulation.

\subsection{Task Planning with POMDP}

Perceived incompleteness emphasizes the necessity of decision-making for efficient completion of manipulation within a task-specific framework. It is not only reflected in the acquisition of complete information, but also in the intensive analysis of task requirements. Such problems can be modeled as POMDP-based approaches. Different from motion planning that considers the spatial movement of the manipulator \cite{kim2016planning} \cite{garg2019learning}, this type of task planning instructs robot the manipulated objects and methods. It is a high-level manipulation indicator and vitally dependent on task settings \cite{lauri2022partially}. Pajarinen et al. \cite{pajarinen2017robotic} design to grasp objects which may be occluded with special attributes and the occlusion information is estimated for planning the best action to be performed. Li et al. \cite{li2016act} direct at searching for objects in a refrigerator, plan a sequence of actions to rearrange objects, and find the target. Xiao et al. \cite{xiao2019online} consider object search for fully occluded objects, use parameterized action to deploy manipulation. Recent works \cite{invigorate} \cite{yang2022interactive} introduce human-robot interaction to facilitate the disambiguation of task instructions. Since the human-specified target is probably unknown or the description language is too vague to confirm the target, POMDP is needed to comprehensively consider observations and human commands to plan action of grasping or asking. Inspired by the above works, we formulate the task as human-like grasping on a dining table, aiming to execute actions according to target requirements without redundancy.

\section{OVERVIEW}

\begin{figure*}[thpb]
  \centering
  \includegraphics[width=17.5cm]{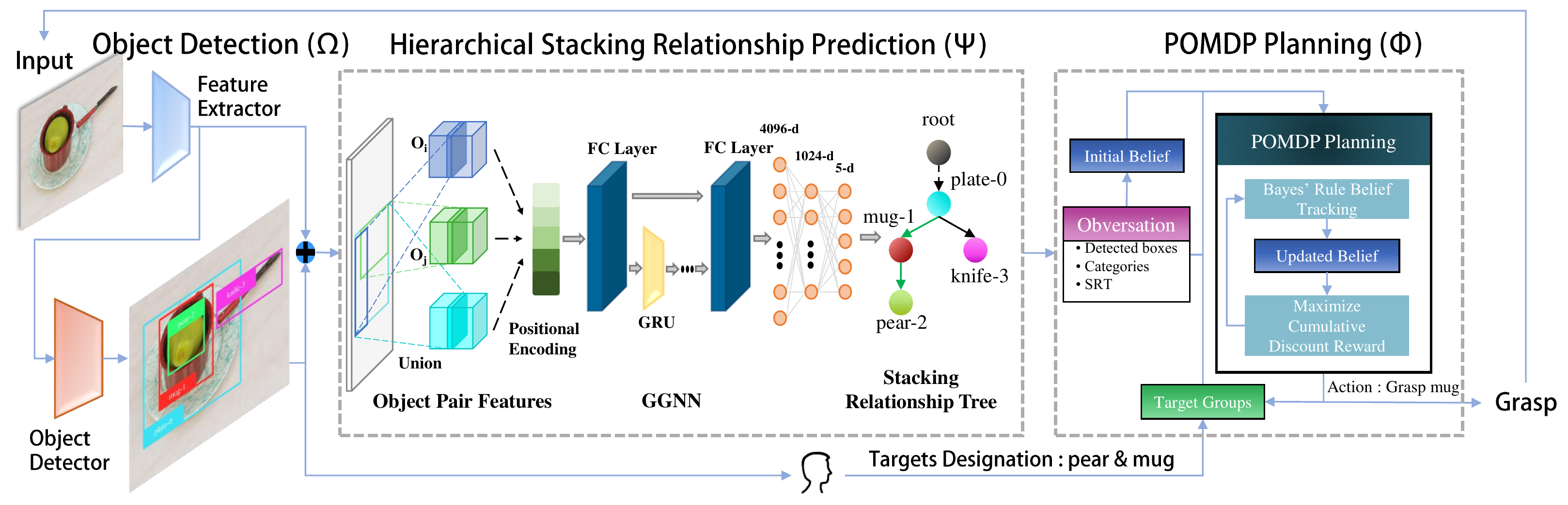}
  \caption{The overview of our proposed approach. The RGB image first needs to go through feature extraction and object detection, then the relationships between object pairs are predicted through the HSRN. Finally, these perception results will be used as observations, and input into the POMDP planner together with human target designation to plan the current optimal grasping action.}
  \label{overviewfig}
\end{figure*}

Our perception-planning architecture is shown in Fig. \ref{overviewfig}, including object detection $\Omega$, relationship prediction $\Psi$, and decision-making planning process $\Phi$. The model incorporates POMDP planning on the deep learning training algorithm. In the training network, we mainly consider object properties and the hierarchical stacking relationship of each object pair. Our model takes RGB images as input, filters a series of proposals for category recognition and bounding-box regression. According to the detected objects, relationship predictor $\Psi$ then constructs the full permutation of all binary object groups to classify their stacking parent-child relationships which are distinguished by stable support and weak support.

The training results are limited by noise, object occlusion, and dataset size. Thus in order to realize robust grasping execution of the robot on the observation results obtained by scene state, we introduce the POMDP-based planning model. POMDP groups targets designated by humans. For a single target group, it assigns action values to all potential actions and updates the belief, which includes historical optimal state value, providing criteria for the selection of each decision-making step. During the planning process, we chiefly consider whether to grasp certain objects together to target area or non-target area, recursively obtain all possible task execution trajectories and at last search out the optimal current grasping action with the highest expected cumulative reward.

\section{PROPOSED APPROACH}

\subsection{Feature Extraction and Object Detection}

The RGB image $I$ captured by the camera first needs to go through a series of convolutional layers for feature extraction, as the input to generate the candidates for target positioning and recognition. Our work continues the target detection proposals in \cite{VMRN}, using ResNet101 \cite{he2016deep} as the feature extractor and Faster R-CNN \cite{ren2015faster} as the object detector. The results of feature, object categories, and bounding boxes will be sent to the relationship prediction network for further analysis.

\subsection{Stacking Relationship Hierarchy Extension}\label{rel extension}

The stacking of objects normally hinders the safety and feasibility of grasping, but some stable stacking can also improve grasping efficiency. Zhang et al. \cite{VMRN} define three stacking relationships between objects. If moving object $o_a$ will change the spatial state and stability of object $o_b$, i.e. $o_a$ supports $o_b$, we call $o_a$ is the parent of $o_b$, on the contrary, $o_a$ is called the child of $o_b$. This kind of relationship classification is not enough to cover the physical spatial relationship and is not adequate to describe scenes with stable stacking. On this basis, we refer to the definition of the support relationship in \cite{panda2013learning} and give:

\begin{itemize}
\item Stable Support: the parent fully supports the child, when the parent is grasped and generates translation in 3d space without rotation, the child will be grasped as well. 
\item Weak Support: the parent partially supports the child, and the child is simultaneously supported by other objects, including the operating table.
\end{itemize}

\begin{figure}[thpb]
  \centering
  \centering
  \subfigure[]{
  \includegraphics[width=7cm]{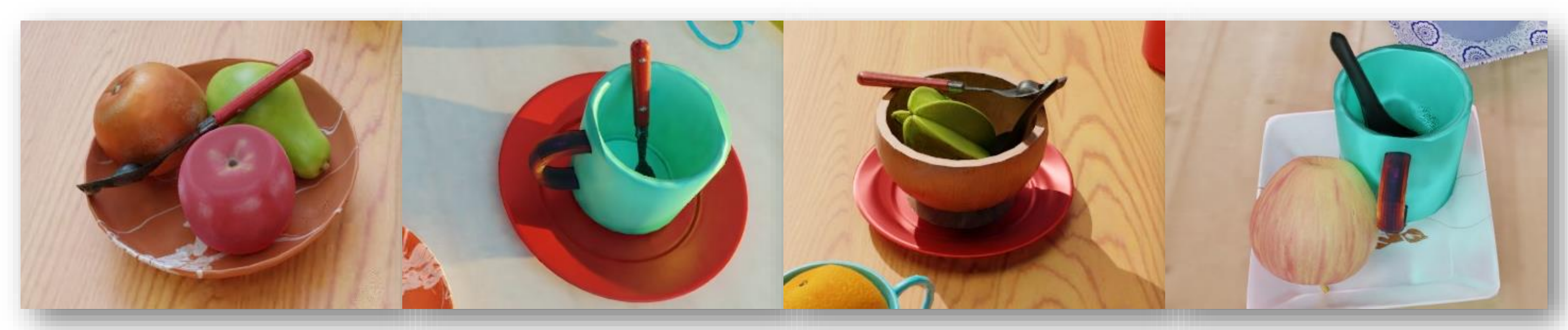}}
  \quad
  \subfigure[]{
  \includegraphics[width=7cm]{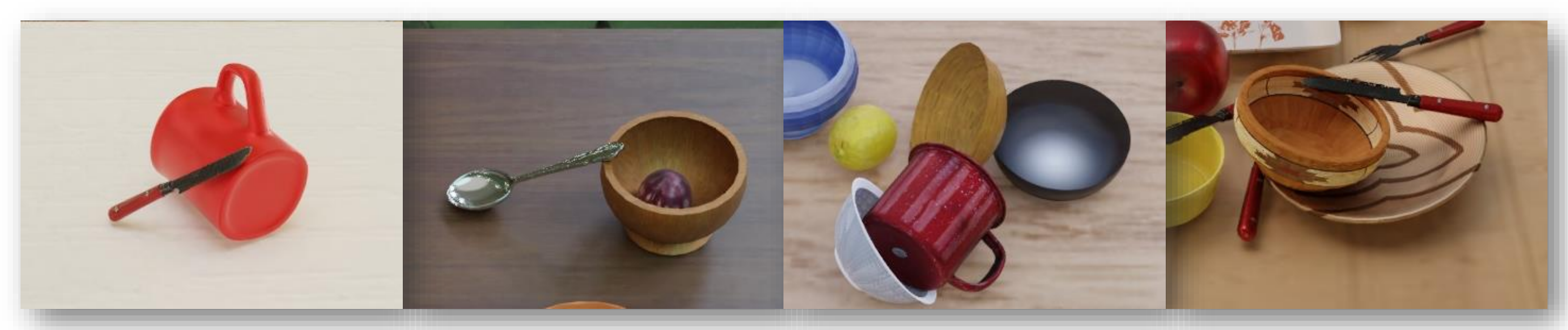}}
  \caption{Different hierarchies of stacking. ($\textit{a}$) Stable Support. ($\textit{b}$) Weak Support.}
  \label{samples_stable}
\end{figure}

Fig. \ref{samples_stable} presents some cases of Stable Support and Weak Support. In Stable Support relationships, the parents generally have container attributes, like bowls. However, in another critical situation although the parent fully supports the child, due to the limits of material rigidity, friction coefficient, and robot manipulation ability, grasping the parent with its child is still difficult, e.g. a book with a pen on it. Thus in the decision-making process, we put forward a thorough planning scheme to decide whether the parent object and child object should be grasped together. 

\subsection{Relationship Prediction}

\begin{figure}[t]
  \centering
  \centering
  \subfigure[]{
  \includegraphics[width=3.5cm]{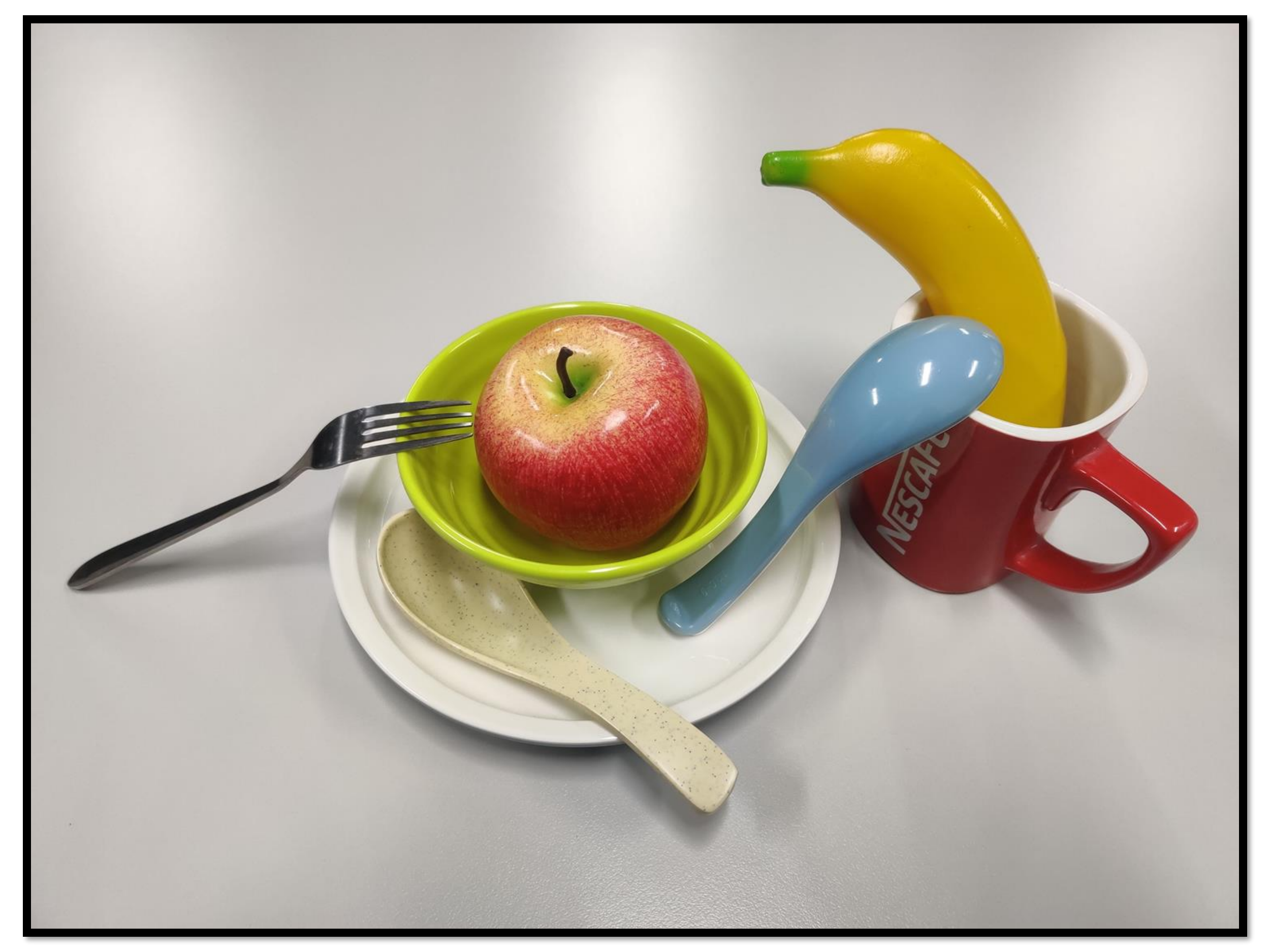}}
  \quad
  \subfigure[]{
  \includegraphics[width=2.8cm]{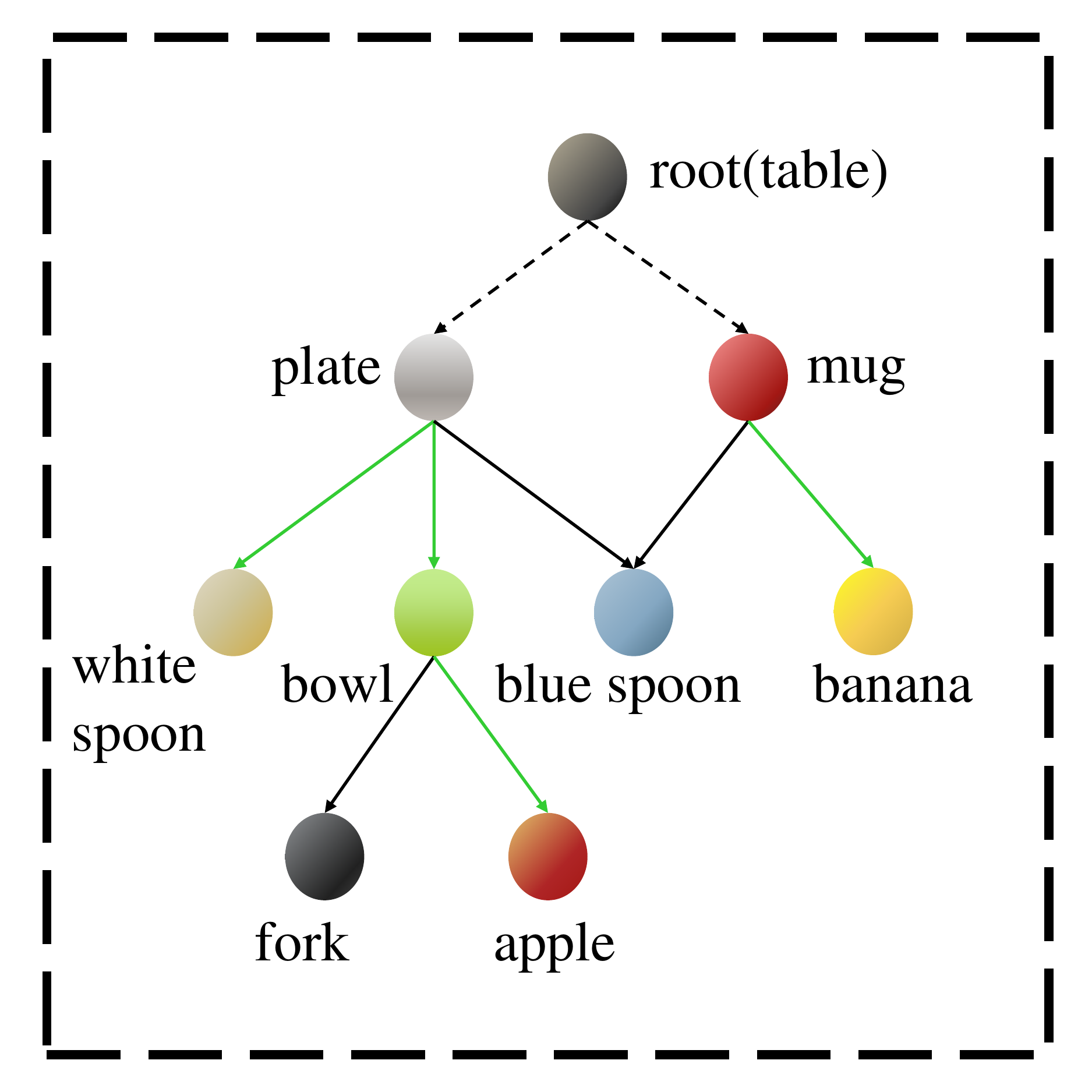}}
  \quad
  \subfigure[]{
  \includegraphics[width=8.6cm]{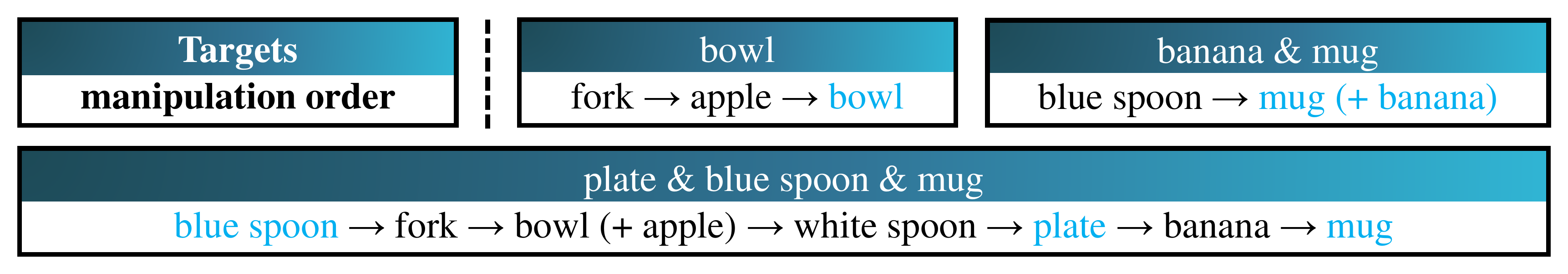}}
  \caption{Relationship prediction and planning results according to specified targets. ($\textit{a}$) The input RGB image. ($\textit{b}$) Stacking Relationship Tree (SRT) to present relationship prediction. The green edge represents the parent node and the child node are stacked in Stable Support, and the black edge represents Weak Support. ($\textit{c}$) The optimal manipulation order chain under the specified targets. The object in parentheses means it will be grasped together. The object in black means it needs to be grasped to the non-target area, and the object in blue is going to be grasped to the target area.}
  \label{all description}
\end{figure}

To predict the relationship properties of each object, our relationship prediction network $\Psi$ extracts the individual and union features of all pairwise combinations of objects $\{\left. \left(o_i, o_j\right)\right|i,j \in N_{obj}\}$ detected by object detector as nodes and apply the Gated Graph Neural Network (GGNN) \cite{ggnn-vmrn} for detecting relationships. We embed the positional encoding into each object-pair node feature, integrate all of the feature node information through Gate Recurrent Unit (GRU) in a full connection manner, and update the hidden-layer vector $h_i^t$ describing node information. We concatenate the GRU output features with initial hidden-layer features to avoid forgetting. Finally, three linear layers are applied to classify relationships, followed by refined SRT representing the prediction results as shown in Fig. \ref{all description}($\textit{b}$). The edges between parent and child nodes are distinguished by stable and weak. In order to standardize, we add a root node to join all the relationship trees, which can be regarded as the operation table. According to the description in Section \ref{rel extension}, classified relationship types $R_c$ can be expressed as follows:

\begin{itemize}
\item $o_a$ is ordinary parent $\left(op\right)$ of $o_b$ (Weak Support)
\item $o_a$ is ordinary child $\left(oc\right)$ of $o_b$ (Weak Support)
\item $o_a$ is natural parent $\left(np\right)$ of $o_b$ (Stable Support)
\item $o_a$ is natural child $\left(nc\right)$ of $o_b$ (Stable Support)
\item $o_a$ and $o_b$ have no support relationship
\end{itemize}

The network uses a multi-class cross-entropy function as the loss function for stacking relationship prediction:

\begin{small}
\begin{equation}
	\begin{aligned}
L_{RP}\left(R ;\Psi \right) & = -\frac{1}{N_{obj}\left(N_{obj}-1\right)} \\
& {\sum\limits_{\left(i,j\right)\in N_{obj}^2, i\ne j}{\sum\limits_{c=1}^5{r_c\mathit{\log}\left({p\left({\left.r_c\right|o_{i},o_{j};\Psi} \right)}\right)}}}
	\end{aligned}	
\end{equation}
\end{small}
and the total loss of our complete network is:
\begin{equation}
	\begin{aligned}
		L\left(I ;\Omega,\Psi \right) = \mu L_{OD}\left(O ;\Omega \right)  + \left( 1-\mu\right) L_{RP}\left(R ;\Psi \right)
	\end{aligned}	
\end{equation}
where $L_{OD}\left(O ;\Omega \right)$ is loss function of object detector $\Omega$, as discussed in \cite{girshick2015fast}. We set the balance weight $\mu$ to 0.5, considering the trade-off with two network modules.

\subsection{POMDP-based Planning}

A POMDP is modeled as a tuple $\left(\mathcal{S}, \mathcal{A}, \mathcal{Z}, \mathcal{T}, \mathcal{O}, \mathcal{R}\right)$, where $\mathcal{S}$, $\mathcal{A}$, $\mathcal{Z}$ respectively denote the set of states, actions, and observations. Transition function $\mathcal{T}\left(s, a,s^{\prime}\right) =p\left(s^{\prime} \mid s, a\right)$ indicates taking action $a \in \mathcal{A}$, the probability from state $s\in \mathcal{S}$ to the next state $s^{\prime}\in \mathcal{S}$. Observation function $\mathcal{O}\left(s^{\prime}, a, z\right)=p\left(z \mid s^{\prime}, a\right)$ is the probability of observing $z \in \mathcal{Z}$ by performing action $a$ and reach the resulting state $s^{\prime}$. $R\left(s, a\right)$ represents the immediate reward of action $a$ in state $a$. What follows are detailed descriptions of each part.

\subsubsection{State}

We factorize the scene state $\mathcal{S}$ to the state of each object $s_i\in \mathcal{S}$ \cite{diuk2008object}. $s_i$ is represented by the Boolean quantity $s^g_i$ of whether it is present in the image (or is grasped and removed) and the relationship between it and other surrounding objects $\left\{s^r_{ij}|j=1, 2,..., N_{obj}\right\}$, the state quantity of $o_i$ is $N_{obj}+1$. The relationship state space can be provided from $R_c$, attached to a value indicating the nonexistence of $s^r_{ii}$, or when $o_i$ is taken away from scenes. Since the real state is unknown due to the partial observation and noise, for each state $s\in \mathcal{S}$, we maintain a belief $b\in \mathcal{B}$ as a state observer to represent the distribution estimation of state. For each object $o_i$, all its beliefs can be expressed as: 
\begin{equation}
\mathcal{B}_i= \left\{b^g_i\right\} \cup \left\{b^r_{ij}|j=1, 2,...,N\right\}\label{B}
\end{equation}
where $b^g_i$ refers to the probability that $o_i$ is observed in the image, and $b^r_{ij}$ refers to the observation relationship between $o_i$ and other objects. We update $b$ in real time after each step of decision-making, re-evaluating the state distribution of the current environment.

\subsubsection{Action}

The task in our work is target-oriented, which means human designates the required targets according to the result of object detection, then send it to POMDP for decision-making. As taking grasping an object an action, the action space is $\left\{2N_{obj} +1\right\}$, including grasping each object to target area or non-target area, alongside a report action that no longer exists grasping action to perform. 

\subsubsection{Observation Model}

We take the results of object detector $\Omega$ and stacking relationship predictor $\Psi$ in HSRN as observations, indicating as $Z^g$ and $Z^r$. The observation function $\mathcal{O} \left(s^{\prime}, a, z\right)$ includes the probability that the updated scene state $s^{\prime}$ can be accurately observed after performing the grasping action $a$, expresses the difference between the real state and the robot perception of the scene. Therefore we learn the probability distribution of observation from the average recall rate index of each object category ${rc}_i \in {RC}$ of both network modules, as it approximates $p\left(z_i \mid s_i\right)$ while $o_i$ is presented in the scene. Since an action $a$ potentially not only causes one object to be grasped, the observation function is comprehensively denoted as $\prod_{j \in C(o_i)}{rc}_j$, where $C(o_i)$ is the set of $o_i$ and its child nodes.

\subsubsection{Transition Model} \label{transition_model}

\begin{figure}[t]
  \centering
  \includegraphics[width=7cm]{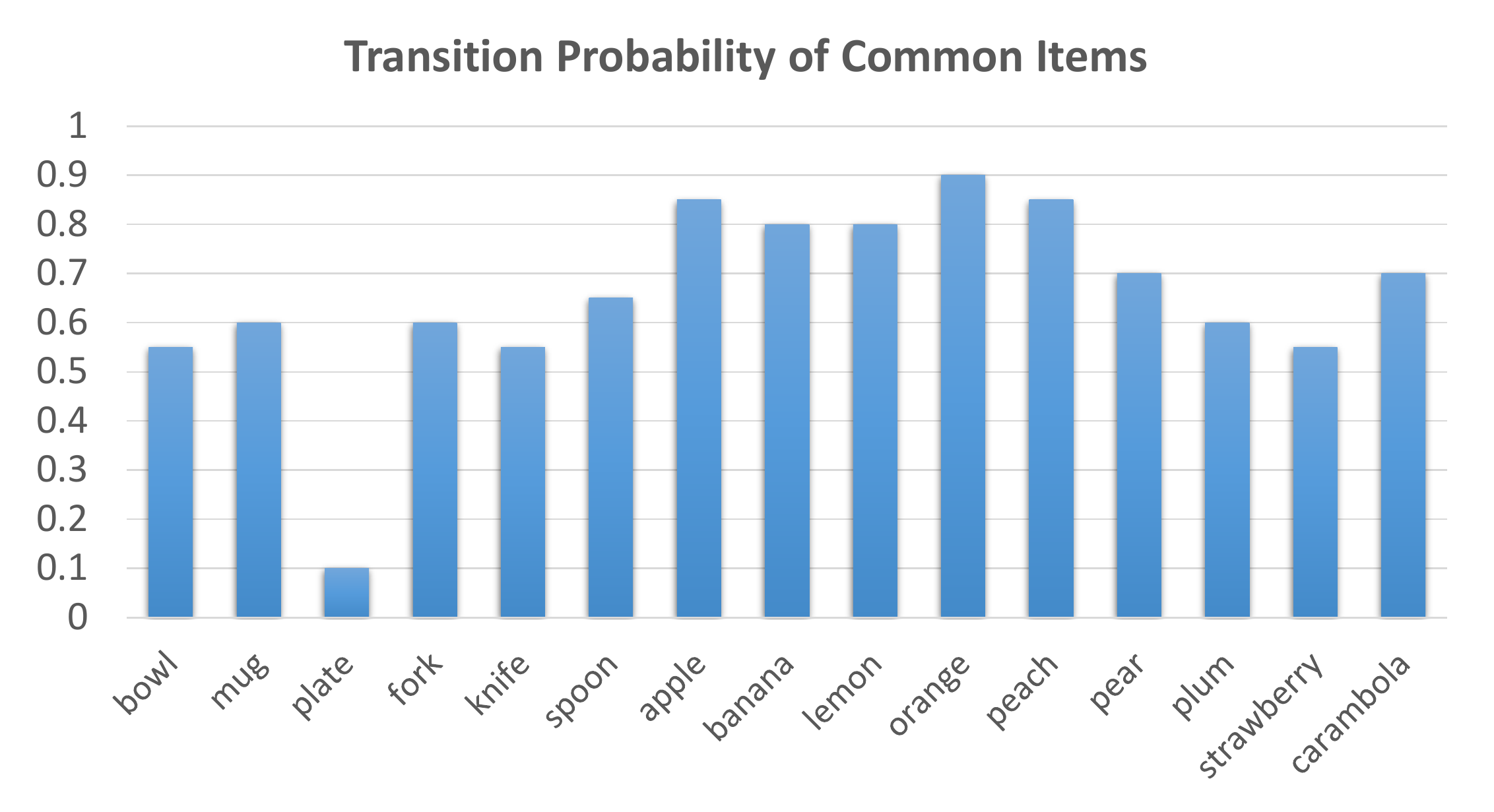} 
  \caption{Transition probability of common object categories.}
  \label{transition}
\end{figure}

Ideally, a grasping action will lead to a state update where the object is no longer presented and its relationship with other objects becomes non-existent. Nevertheless, the robot's manipulative capacity results in different performances in grasping various types of objects. Therefore we collect a series of empirical data in our robot experimental environment about the capacity to grasp types of objects separately. We choose 15 common item categories in the dining table scenes, for each category, measure the success rate that it can be grasped by the gripper horizontally and stably by performing grasping 20 times. The grasping method \cite{roi-grasp} is adopted for our data collection. We quantify the average success rate with data normalization to obtain the transition probability $p\left(s^{\prime} \mid s, a\right)$ of each object (see Fig. \ref{transition}). If grasp fails, we default the state does not change to simplify decision-making reasoning. In experiments, we revise the data based on the real grasping success rate and verify the validity of the transition modeling.

\subsubsection{Reward Function}

We encourage the robot to complete the task with the least number of grasps, so we give the agent a base penalty of $-10$ for each grasp action. The relationship between the executed object and the other will generate additional rewards, we empirically design a relation-based grasping reward for each object:
\begin{small}
\begin{equation}
R\left(s, a_{i}\right) = -10 +
\begin{cases} 
5\tanh{\left(N_{nc}\left( o_{i} \right)\right)},  & \text{if } N_{nc}\left( o_{i} \right)>0\\
-10\tanh{\left(N_{oc}\left( o_{i} \right)\right)}, & \text{if }N_{oc}\left( o_{i} \right)>0 \\
-2, & \text{if }N_{np}\left( o_{i} \right)>0 \\
0, & \text{else} \\
\end{cases}
\end{equation}
\end{small}
where $N_{nc}$, $N_{oc}$, and $N_{np}$ describe the number of natural children, ordinary children, and natural parents of each object. In an SRT, if $o_{i}$ has natural children, we expect it to be grasped actively and give it a major reward. Hyperbolic tangent function $\tanh \left(x\right)=\frac{e^x-e^{-x}}{e^x+e^{-x}}$ fixes reward upper limit. When $o_{i}$ has ordinary children, it is unwarrantable to be grasped by giving a large penalty. If $o_{i}$ has a natural parent, it is in principle not encouraged to be grasped directly, with a slight penalty of $-2$. If $o_{i}$ is a leaf or isolated node, no reward will be attached. Since we set the scene state value when the task is completed to 0, the purpose of this design is to avoid the optimal state value obtained at each step being greater than 0.

\subsubsection{Belief Updating $\&$ Planning}

In partially observable domains, after taking action $a$, the observation $z$ is received and belief $b$ will be updated to $b^{\prime}$. As mentioned in \eqref{B}, we refine the scene state to the private state of each object and update each belief individually. The planner starts with an initial belief $b_0$, applies Bayes' rule to track the belief at each step. Specifically:
\begin{equation}
b^{\prime}\left(s^{\prime}\right)=\lambda \mathcal{O}\left(s^{\prime}, a, z\right) \sum_s \mathcal{T}\left(s, a, s^{\prime}\right) b(s)
\end{equation}
where $\lambda=\frac{1}{p\left(o \mid a,b\right)}$ is a Bayes' rule's constant for normalization.

For the SRT generated by $\Psi$, we cannot directly reference it for planning, after all our purpose is not just cleaning the table (of course it also can be), but to grasp the desired targets. To this end, we first construct the Descendant Hash Table (DHT) of each object according to SRT. In the table, for each object $o_i$, all objects in the subtree with $o_i$ as the root are arranged in order from leaf to root, the last element is $o_i$ itself, i.e. the action space when $o_i$ is the target. 

It has been reflected in section \ref{rel extension} that parent-child nodes connected by Stable Support are not necessarily grasped together. After identifying the corresponding prediction results and obtaining the targets designation $D$, the planning process first needs to group the targets, because stacking relationships may exist between the targets themselves. Our planner incorporates the target nodes connected by the Stable Support relationship into a target group $g \in G$. Since targets and non-targets should not be grasped together, it downgrades Stable Support between target groups and non-targets to Weak. Then the planner sort all groups from leaf to root, sequentially search all descendants of each group based on DHT as action, and implement planning. After all nodes in the current action space are pruned, switch to the next group till the end. In planning process, a policy $\pi:\mathcal{B} \rightarrow\mathcal{A}$ is learned to maximize the cumulative discount reward according to the initial belief $b_0$:
\begin{equation}
V^\pi\left(b_0\right)=\mathbb{E}\left[\sum_{t=0}^{\infty} \gamma^t \mathcal{R}\left(s_t, a_t\right) \bigg| a_t=\pi\left(b_t\right)\right]
\end{equation}
where the discount factor $\gamma$ is set to 0.8. Last, carry out the look-ahead search to find the optimal action:
\begin{equation}
a^*=\arg \max _a V^\pi\left(b\right)
\end{equation}

We concatenate the grasp action of each step in the planning process into a chain as a manipulation decision chain, as presented in Fig. \ref{all description}($\textit{c}$).

\section{Experiment}
\subsection{Dataset Construction}

\begin{figure}[t]
  \centering
  \centering
  \includegraphics[width=8.5cm]{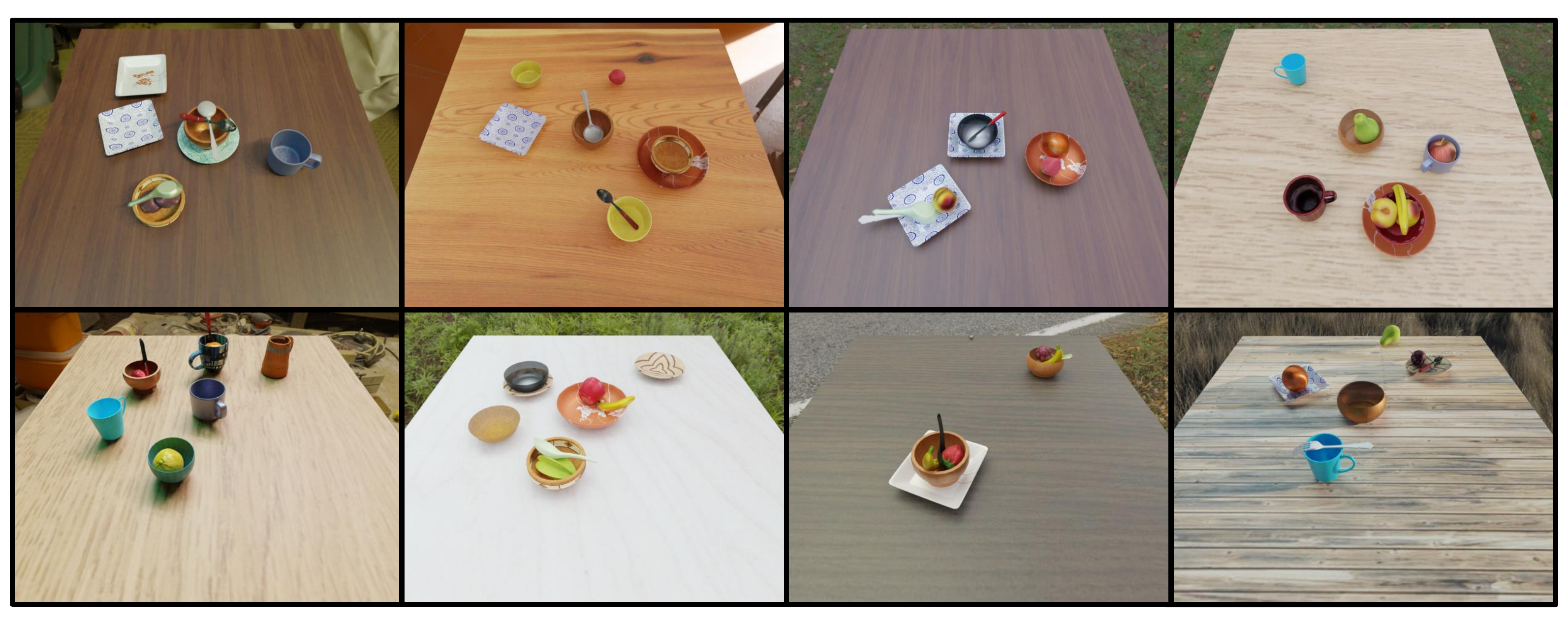}
  \caption{Some examples of REGRAD-v2. Objects are mostly stacked in order, fully simulating the stacking method on real dining tables.}
  \label{dataset}
\end{figure}

On the basis of REGRAD \cite{regrad}, we expand the dining scene dataset REGRAD-v2 in a virtual environment, replenish 15 object categories of main tableware objects (plate, bowl, mug, spoon, fork, knife) and some fruits (apple, banana, etc.). The models with an amount of 44 come from Shapenet \cite{chang2015shapenet}, and YCB dataset \cite{calli2017yale}, all dimensions are based on real objects. In order to prevent model penetration due to the initial z-axis height limit, we use Poisson disk sampling \cite{bridson2007fast} to uniformly select the plane position of model loading, load other tableware and fruits after the container objects are placed. Each scene contains 8 to 12 objects. In addition, we also generate substantial cluttered scenes in REGRAD standard procedure to supplement negative samples. Our newly generated dataset contains 3.2 k scenes and entirely shares all properties and functions with the original dataset. On the basis of the automatic label generation method \cite{regrad} of object bounding box, object category, 2D grasping position, and manipulation relationship (not distinguish Stable and Weak), we manually label all Stable Support relationships. Some scenes in REGRAD-v2 are shown in Fig. \ref{dataset}.

\subsection{Perception and Decision}

\subsubsection{Implementation Details} \label{imp details}

Our model is implemented in the PyTorch framework and uses an NVIDIA RTX 3090 with 24GB of memory to train, the maximum training epoch is 30. The object detector used is Faster R-CNN \cite{ren2015faster} with the backbone of pre-trained ResNet101 \cite{he2016deep}. We set the learning rate to 0.001 and decayed to 0.0001 after 10k iterations. Stochastic Gradient Descent(SGD) with a momentum of 0.9 is used as the optimizer. All hyperparameters are shared during training.

For each scene in REGRAD-v2, we collect 2 groups of artificially designated single target and 2 groups of multiple targets as task requirements, with an extra task of table cleaning in which all objects are regarded as targets. We specifically annotate the desired grasping decision chain by human anticipation according to the principle in section \ref{intro}. and utilize perception results for planning, in order to verify the effectiveness and rationality of our POMDP model. In the generated chain, for some unrelated objects, the order of grasping does not affect the rationality of the overall decision-making. So we are concerned about whether the robot plans redundant or omits necessary grasping actions rather than pursuing complete alignment with annotated decision chain. 

\subsubsection{Evaluation Metrics}

The metrics of the perception process continue the baseline of previous work for visual manipulation relationship detection \cite{VMRN}. On this basis, we design the measurement standard of the planning process.

\begin{itemize}
\item Object detection: \textbf{mAP} (main Average Precision) measures the average precision of all object categories, which is commonly used in object recognition.
\item Relationship prediction: The metrics used to evaluate predicted relationship indicated as triplet $<o_i,r_{ij},o_j>$ are \textbf{OR} (Object triplet Recall), \textbf{OP} (Object triplet Precision) and \textbf{IA} (Image-Wise Accuracy). \textbf{IA} calculates the proportion of all objects and triple relationships are accurately predicted in an image.   
\item Decision Making: Planning rationality criterion is discussed in Section \ref{imp details}. We refer to the set of all objects in a decision chain as Action Object Set (AOS). In particular, if AOS generated by the planning process is not equal to the AOS of annotated chain, i.e. the grasping actions are redundant or missing, the decision chain is irrational. Since we update the scene and re-decision after each action to ensure robustness, we evaluate the average rationality of the first step decision $AR_f$ and the whole decision chain $AR_w$ separately. 
\end{itemize}

\subsubsection{Performance}

\begin{figure}[t]
  \label{results_pic}
  \centering
  \includegraphics[width=7.5cm]{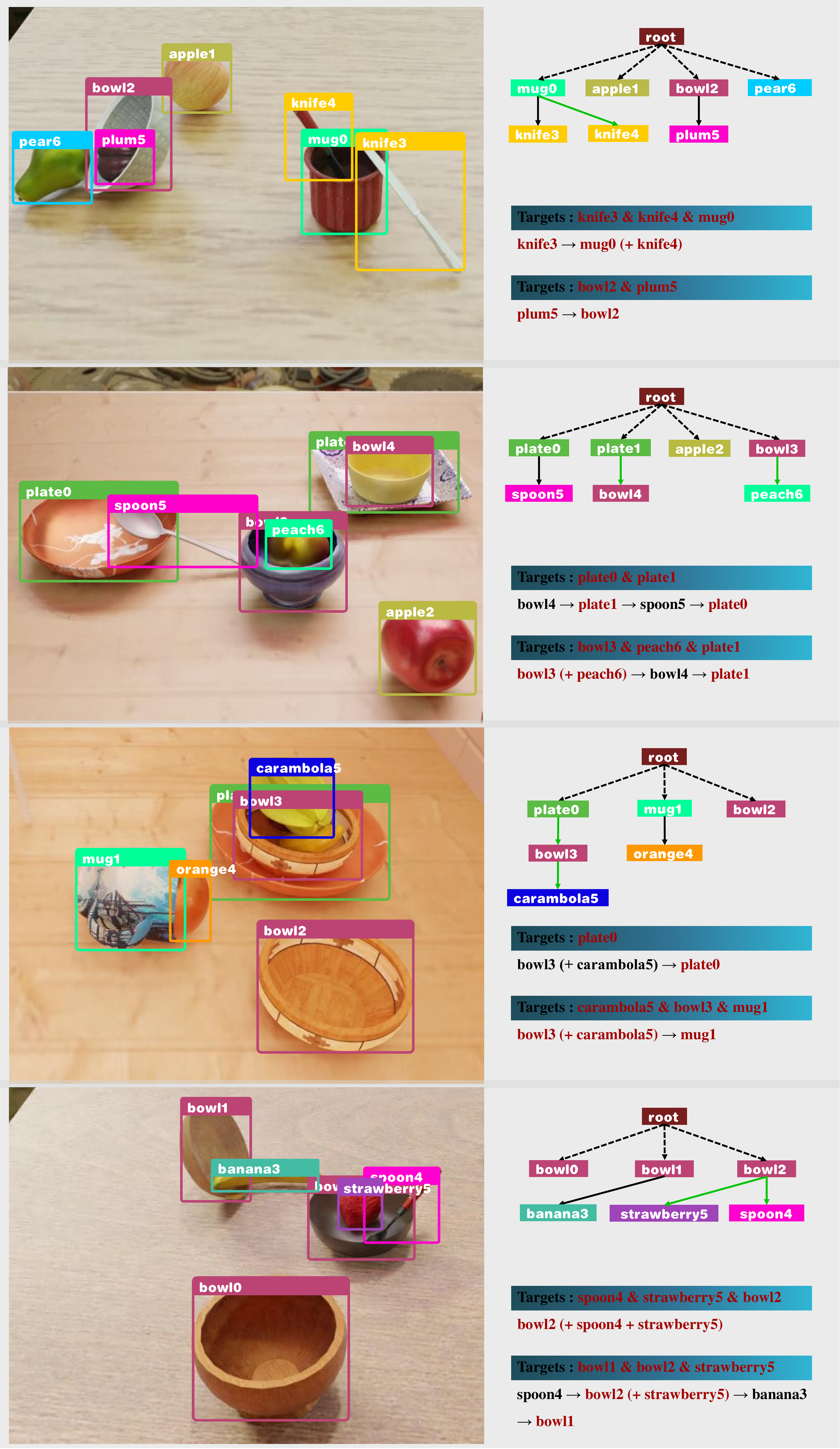}
  \caption{Perception and planning results of our method. \textbf{Left} Object detection results. \textbf{Top right} Stacking Relationship prediction results that presented by SRT. \textbf{Bottom right} Planning grasp-decision chain based on human-given target designations.}
  \label{per-pla results}
\end{figure}

Our perception-planning results are shown in Fig. \ref{per-pla results}. We compare our method with state-of-the-art stacking relationship detection baselines performed on the REGRAD-v2 dataset. Implementation settings are the same as our method, using ResNet101 for feature extraction and Faster R-CNN for object detection. Relationship prediction in these baselines only considers original parent-child relationships defined by \cite{VMRN}. Table \ref{img-acc} shows the results that our algorithm is better than other baselines generally. Although the hierarchy of relationships is more complex in our work, the dining table scenes have more restrictions on the representation of object stacking than clutter scenes, which facilitates our network to learn the staking modes of objects. Furthermore, GGNN integrates context information, fully utilizes global scene description to comprehend Stable and Weak Support, ensuring comprehensive perception performance. 

\begin{table}[t]
\renewcommand\arraystretch{1.4}
\caption{PERFORMANCE OF OBJECT DETECTION AND RELATIONSHIP PREDICTION BASED ON REGRAD-V2 DATESET}
\label{img-acc}
\begin{center}
\begin{tabular}{c|cccc}
\hline
\makebox[0.1\textwidth][c]{\multirow{2}{*}{\bf Algorithm}} 
& \multicolumn{4}{c}{\bf Metrics ($\%$)} \\\cline{2-5}
& \makebox[0.06\textwidth][c]{\bf mAP} 
& \makebox[0.06\textwidth][c]{\bf OR} 
& \makebox[0.06\textwidth][c]{\bf OP}  
& \makebox[0.06\textwidth][c]{\bf IA} \\
\hline
VMRN\cite{VMRN} & 90.28 & 66.78 & 65.82 & 21.40 \\
GVMRN-RF\cite{zuo_graph} & 89.90 & 68.59 & 68.31 & 24.06  \\
GGNN-VMRN\cite{ggnn-vmrn} & 90.54 & 72.26 & 71.95 &  26.62 \\
\bf HSRN(ours) & \textbf{90.69} & \textbf{72.84} & \textbf{73.53} & \textbf{26.85} \\
\hline
\end{tabular}
\end{center}
\end{table}

\begin{table}[t]
\renewcommand\arraystretch{1.4}
\caption{RATIONALITY ($\%$) OF PLANNING ACTION UNDER DIFFERENT TASK SETTINGS}
\label{rationlity}
\begin{center}
\begin{tabular}{c|cc|cc|cc}
\hline
\makebox[0.038\textwidth][c]{\multirow{2}{*}{\bf Algorithm}} 
& \multicolumn{2}{c|}{\bf \textit{Task ST}} & \multicolumn{2}{c|}{\bf \textit{Task MT}}& \multicolumn{2}{c}{\bf \textit{Task TC}}\\\cline{2-7}
& \makebox[0.028\textwidth][c]{\bf $AR_f$} 
& \makebox[0.028\textwidth][c]{\bf $AR_w$} 
& \makebox[0.028\textwidth][c]{\bf $AR_f$}  
& \makebox[0.028\textwidth][c]{\bf $AR_w$} 
& \makebox[0.028\textwidth][c]{\bf $AR_f$}  
& \makebox[0.028\textwidth][c]{\bf $AR_w$} \\
\hline
VMRN\cite{VMRN} & 97.65 & 96.54 & 62.40 & 47.36 & 76.42 & 24.88\\
\bf HSRN & 92.30 & 87.55 & 75.02 & 58.59 & 96.46 & \textbf{90.81}  \\
\bf HSRN-POMDP & \textbf{98.64} & \textbf{97.37} & \textbf{98.25} & \textbf{95.63} & \textbf{97.31} & 86.52 \\
\hline
\end{tabular}
\end{center}
\end{table}

The implementation of POMDP planning will reduce unnecessary grasping steps and make robotic action more in line with human habits. Since existing baselines \cite{VMRN} \cite{zuo_graph} \cite{ggnn-vmrn} specify stacking relationship as the manipulation relationship, when the relationship is predicted, they can directly search action decision. The search method is all the same with these baselines, which is subsequent traversing from SRT. Thus we only compare the average rationality $AR_f$ and $AR_w$ of decision-making with \cite{VMRN} baseline, as shown in Table \ref{rationlity}. Task \textit{ST}, \textit{MT}, \textit{TC} respectively means single target grasping task, multiple targets grasping task, and table cleaning task. 

It is clear to see that due to the lack of hierarchical distinction between the relationship types, \cite{VMRN} can only grasp objects one by one, which is irrational especially in task \textit{MT} and \textit{TC} on account of planning redundant grasps. We also give the decision results that do not go through the POMDP planning, only perform action relying on HSRN and the principle in Section \ref{intro} that if nodes are connected with Stable Support, grasp the parent node. Apparently, it is arbitrary to plan in this way, amounts of necessary grasps are ignored. It is likely to perform the action which grasping targets with non-targets into the target area together. With the implementation of the POMDP decision-making process, the planning chain with high rationality can be obtained for all three tasks. The POMDP also reflects some compromises made in the design of the robot's capabilities which is unfortunately against human hope. For example, the difficulty of stably grasping the plate is too high, so when the plate has a natural child such as an apple, they will still be grasped separately. So in task \textit{TC}, $AR_w$ of HSRN-POMDP is not as high as directly plan from HSRN. This is the result of considering comprehensive factors.

\subsection{Robot Manipulation}

\subsubsection{Experimental Setup}

\begin{figure}[t]
  \centering
  \subfigure[]{
  \includegraphics[width=2.9cm]{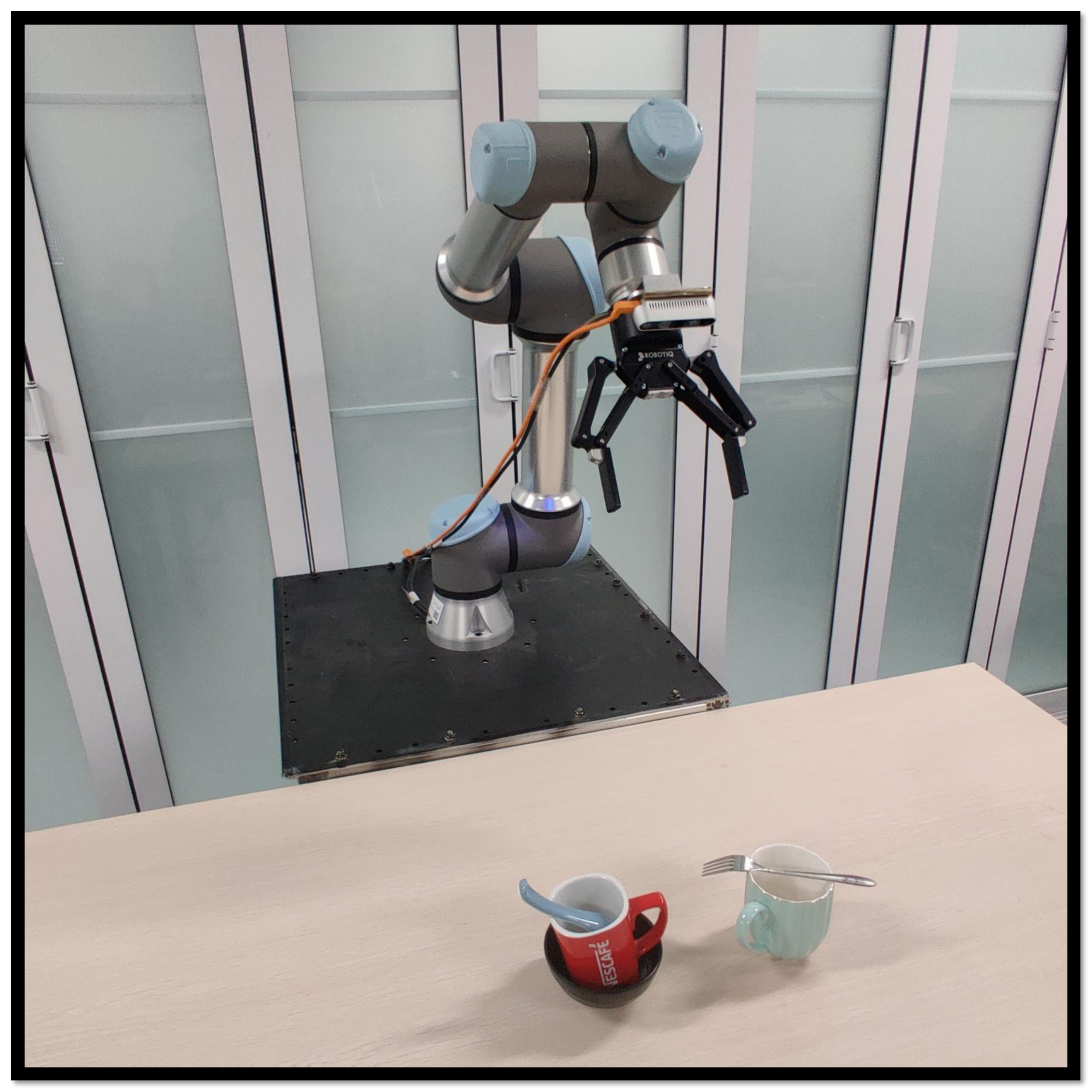}}
  \quad
  \subfigure[]{
  \includegraphics[width=4.3cm]{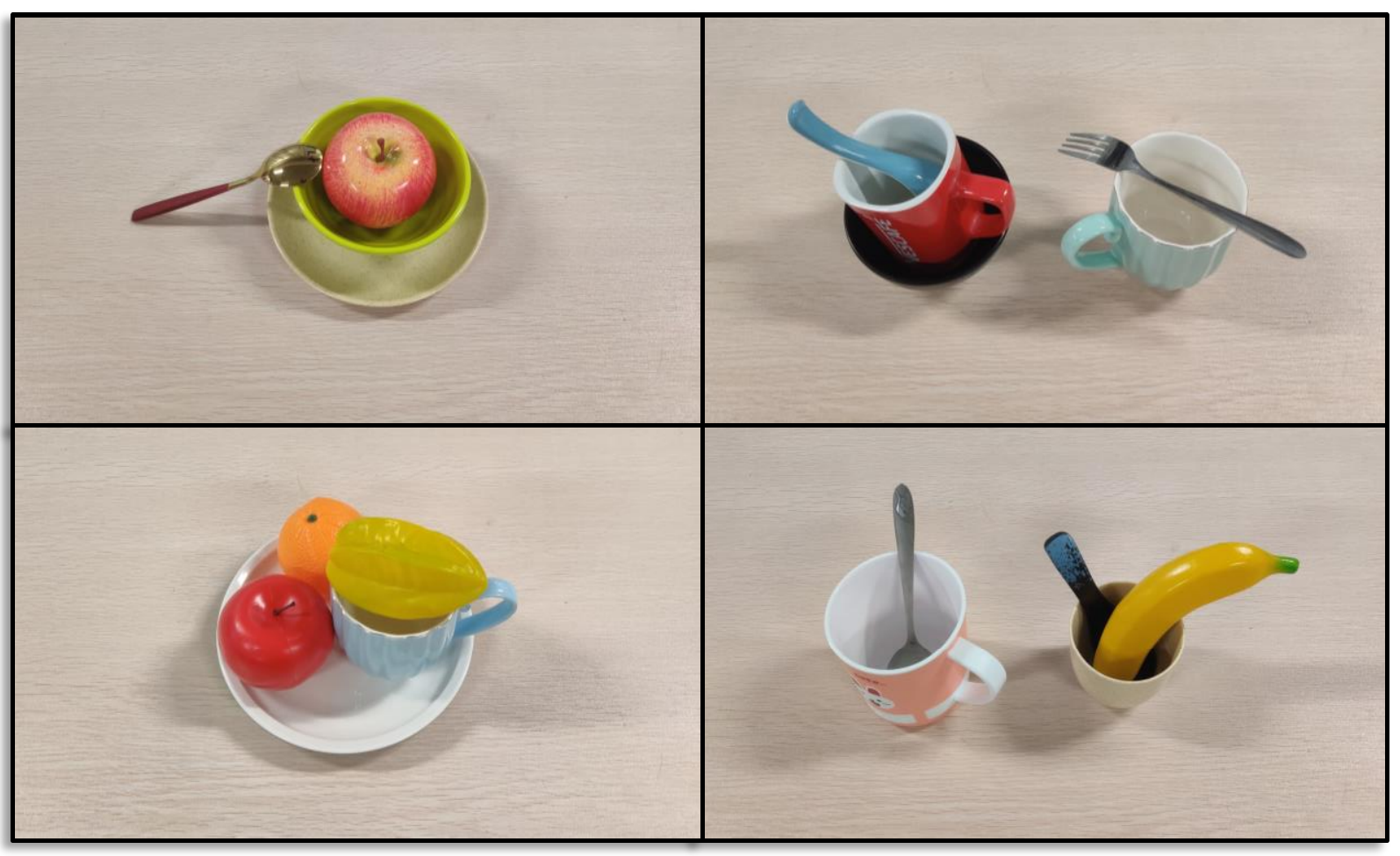}}
  \caption{Robot manipulation configuration. ($\textit{a}$) Experimental environment. ($\textit{b}$) Examples of setting scenes for grasping tasks.}
  \label{real_exp_scene}
\end{figure}

We use a UR5e robot to perform our grasping experiments. The robot is equipped with a 2-finger Robotiq 2F-140 gripper and an eye-in-hand Intel RealSense D435i camera as shown in Fig. \ref{real_exp_scene}($\textit{a}$). We use ROS to drive the robot and gripper. The real models used include bowls, mugs, tableware, and fruits, that are stacked on the operating table in an orderly manner, as shown in Fig. \ref{real_exp_scene}. After the robot perceives the scene and obtains the detection result, it informs humans of the detected object categories with serial numbers to ensure the uniqueness of each object label. The human then specifies the desired targets and the robot plans to grasp them to complete the task. 

\begin{table}[t]
\renewcommand\arraystretch{1.4}
\caption{SUCCESS RATE AND EXECUTION EFFICIENCY OF ROBOT MANIPULATION IN DIFFERENT TASKS}
\label{robot experiment}
\begin{center}
\begin{tabular}{c|c|ccc}
\hline
\makebox[0.05\textwidth][c]{\multirow{2}{*}{\bf Task}}
& \makebox[0.09\textwidth][c]{\multirow{2}{*}{\bf Algorithm}}
& \makebox[0.06\textwidth][c]{\bf Success}
& \makebox[0.05\textwidth][c]{\bf Number}
& \makebox[0.06\textwidth][c]{\bf Time Cost} \\
& & {\bf Rate ($\%$)} & {\bf of Grasps} & {\bf ($min$)}\\
\hline
\multirow{2}{*}{\bf \textit{ST}} & GGNN-VMRN\cite{ggnn-vmrn} & 10/12 &2.83&2.63 \\
& \textbf{HSRN-POMDP} & \textbf{11/12} & \textbf{2.08} & \textbf{1.89} \\
\hline
\multirow{2}{*}{\bf \textit{MT}} & GGNN-VMRN\cite{ggnn-vmrn} & 7/12 & 4.67 & 4.24 \\
& \textbf{HSRN-POMDP} & \textbf{10/12} & \textbf{3.52} & \textbf{3.20} \\
\hline
\multirow{2}{*}{\bf \textit{TC}} & GGNN-VMRN\cite{ggnn-vmrn} & 4/12 & 5.66 & 5.17 \\
& \textbf{HSRN-POMDP} & \textbf{7/12} & \textbf{3.58} & \textbf{3.32} \\
\hline
\end{tabular}
\end{center}
\end{table}

We set up 12 scenes in total, with 3$\sim$6 objects in each scene, and each scene is divided into three tasks \textit{ST}, \textit{MT}, and \textit{TC}. In experiments, we restore the initial scene as consistently as possible. The criterion for task success is that all specified objects are grasped and placed in the target area. We compare our method with the baseline of state-of-the-art VMRN-based method \cite{ggnn-vmrn}. We adopt ROI-GD \cite{roi-grasp} method for our grasp position detection, which is as same as used in Section \ref{transition_model}. We train the detection network based on the 2D grasping position generated during dataset construction. In the experiment, we fine-tune the grasping position by discretized searching based on the detected grasping position, select the position whose projection in the image mostly overlaps the detected bounding box. When grasping objects such as bowls and mugs, our two-finger gripper first adjusts to horizontal orientation, and approaches the object in a horizontal manner. When grasping objects such as fruits, the gripper approaches the object in a top-down direction to conform to human grasping habits. 

\subsubsection{Results}

Table \ref{robot experiment} summarizes the success rate, average number of performed grasps, and average time cost of our method with baseline under different task settings. The results reveal that our network generalizes well to real grasping scenarios and performs better than the GGNN-VMRN method in terms of task completion. In experiments, our method saves an average of $29.48\%$ on task execution time, because some grasping actions are legitimately saved while ensuring execution stability, and the entire manipulation process is more in line with human behavior habits.

Due to the limited width of the jaws, when a plate is in targets, it is barely able to be successfully grasped. Comfortingly, POMDP effectively judges the children stacked on plates as a separate grasp, not limited to Stable and Weak relationship predictions. Other error cases show the incompleteness of decision-making. For example, the fork is stably supported by the mug, but it is in a state of lying across the rim of the mug. Our planning process has a considerable probability of directly grasping the mug. Ideally, there would be no problem with this action, but in practice, this may cause the fork to drop during movement.

\section{Conclusion}

This paper proposes a hierarchical object stacking relationship detection network and introduces a POMDP-based decision-making process for giving a rational grasping execution process for specific tasks. Experiments show that our algorithm greatly simplifies the grasping process while ensuring the success rate. Future work will pay more attention to the in-depth understanding of the scene and human requirements, analyzing and reconstructing the scene to complete tasks in a more humanized way. Besides, we will pursue further improvement in the detection efficiency of different types of stacking relationships, and provide incisive robotic understanding for robust grasping in multiple scenarios.



\bibliography{IEEEabrv,ref}

\begin{thebibliography}{10}
\providecommand{\url}[1]{#1}
\csname url@rmstyle\endcsname
\providecommand{\newblock}{\relax}
\providecommand{\bibinfo}[2]{#2}
\providecommand\BIBentrySTDinterwordspacing{\spaceskip=0pt\relax}
\providecommand\BIBentryALTinterwordstretchfactor{4}
\providecommand\BIBentryALTinterwordspacing{\spaceskip=\fontdimen2\font plus
\BIBentryALTinterwordstretchfactor\fontdimen3\font minus
  \fontdimen4\font\relax}
\providecommand\BIBforeignlanguage[2]{{%
\expandafter\ifx\csname l@#1\endcsname\relax
\typeout{** WARNING: IEEEtran.bst: No hyphenation pattern has been}%
\typeout{** loaded for the language `#1'. Using the pattern for}%
\typeout{** the default language instead.}%
\else
\language=\csname l@#1\endcsname
\fi
#2}}

\bibitem{liu2012fast}
M.-Y. Liu, O.~Tuzel, A.~Veeraraghavan, Y.~Taguchi, T.~K. Marks, and
  R.~Chellappa, ``Fast object localization and pose estimation in heavy clutter
  for robotic bin picking,'' \emph{The International Journal of Robotics
  Research}, vol.~31, no.~8, pp. 951--973, 2012.

\bibitem{murali20206}
A.~Murali, A.~Mousavian, C.~Eppner, C.~Paxton, and D.~Fox, ``6-dof grasping for
  target-driven object manipulation in clutter,'' in \emph{2020 IEEE
  International Conference on Robotics and Automation (ICRA)}.\hskip 1em plus
  0.5em minus 0.4em\relax IEEE, 2020, pp. 6232--6238.

\bibitem{fischinger2013learning}
D.~Fischinger, M.~Vincze, and Y.~Jiang, ``Learning grasps for unknown objects
  in cluttered scenes,'' in \emph{2013 IEEE international conference on
  robotics and automation}.\hskip 1em plus 0.5em minus 0.4em\relax IEEE, 2013,
  pp. 609--616.

\bibitem{VMRN}
H.~Zhang, X.~Lan, X.~Zhou, Z.~Tian, Y.~Zhang, and N.~Zheng, ``Visual
  manipulation relationship network for autonomous robotics,'' in \emph{2018
  IEEE-RAS 18th International Conference on Humanoid Robots (Humanoids)}.\hskip
  1em plus 0.5em minus 0.4em\relax IEEE, 2018, pp. 118--125.

\bibitem{ggnn-vmrn}
M.~Ding, Y.~Liu, C.~Yang, and X.~Lan, ``Visual manipulation relationship
  detection based on gated graph neural network for robotic grasping,'' in
  \emph{2022 IEEE/RSJ International Conference on Intelligent Robots and
  Systems (IROS)}.\hskip 1em plus 0.5em minus 0.4em\relax IEEE, 2022, pp.
  1404--1410.

\bibitem{tchuiev2022duqim}
V.~Tchuiev, Y.~Miron, and D.~Di~Castro, ``Duqim-net: Probabilistic object
  hierarchy representation for multi-view manipulation,'' in \emph{2022
  IEEE/RSJ International Conference on Intelligent Robots and Systems
  (IROS)}.\hskip 1em plus 0.5em minus 0.4em\relax IEEE, 2022, pp.
  10\,470--10\,477.

\bibitem{regrad}
H.~Zhang, D.~Yang, H.~Wang, B.~Zhao, X.~Lan, J.~Ding, and N.~Zheng, ``Regrad: A
  large-scale relational grasp dataset for safe and object-specific robotic
  grasping in clutter,'' \emph{IEEE Robotics and Automation Letters}, vol.~7,
  no.~2, pp. 2929--2936, 2022.

\bibitem{farhadi2011recognition}
A.~Farhadi and A.~Sadeghi, ``Recognition using visual phrases,'' in
  \emph{Computer Vision and Pattern Recognition (CVPR)}, 2011.

\bibitem{divvala2014learning}
S.~K. Divvala, A.~Farhadi, and C.~Guestrin, ``Learning everything about
  anything: Webly-supervised visual concept learning,'' in \emph{Proceedings of
  the IEEE Conference on Computer Vision and Pattern Recognition}, 2014, pp.
  3270--3277.

\bibitem{lu2016visual}
C.~Lu, R.~Krishna, M.~Bernstein, and L.~Fei-Fei, ``Visual relationship
  detection with language priors,'' in \emph{Computer Vision--ECCV 2016: 14th
  European Conference, Amsterdam, The Netherlands, October 11--14, 2016,
  Proceedings, Part I 14}.\hskip 1em plus 0.5em minus 0.4em\relax Springer,
  2016, pp. 852--869.

\bibitem{liang2018visual}
K.~Liang, Y.~Guo, H.~Chang, and X.~Chen, ``Visual relationship detection with
  deep structural ranking,'' in \emph{Thirty-Second AAAI Conference on
  Artificial Intelligence}, 2018.

\bibitem{panda2013learning}
S.~Panda, A.~A. Hafez, and C.~Jawahar, ``Learning support order for
  manipulation in clutter,'' in \emph{2013 IEEE/RSJ international conference on
  intelligent robots and systems}.\hskip 1em plus 0.5em minus 0.4em\relax IEEE,
  2013, pp. 809--815.

\bibitem{crf}
C.~Yang, X.~Lan, H.~Zhang, X.~Zhou, and N.~Zheng, ``Visual manipulation
  relationship detection with fully connected crfs for autonomous robotic
  grasp,'' in \emph{2018 IEEE International Conference on Robotics and
  Biomimetics (ROBIO)}.\hskip 1em plus 0.5em minus 0.4em\relax IEEE, 2018, pp.
  393--400.

\bibitem{zuo_graph}
G.~Zuo, J.~Tong, H.~Liu, W.~Chen, and J.~Li, ``Graph-based visual manipulation
  relationship reasoning network for robotic grasping,'' \emph{Frontiers in
  Neurorobotics}, vol.~15, p. 719731, 2021.

\bibitem{zhu2020deformable}
X.~Zhu, W.~Su, L.~Lu, B.~Li, X.~Wang, and J.~Dai, ``Deformable detr: Deformable
  transformers for end-to-end object detection,'' \emph{arXiv preprint
  arXiv:2010.04159}, 2020.

\bibitem{kim2016planning}
S.-K. Kim and M.~Likhachev, ``Planning for grasp selection of partially
  occluded objects,'' in \emph{2016 IEEE International Conference on Robotics
  and Automation (ICRA)}.\hskip 1em plus 0.5em minus 0.4em\relax IEEE, 2016,
  pp. 3971--3978.

\bibitem{garg2019learning}
N.~P. Garg, D.~Hsu, and W.~S. Lee, ``Learning to grasp under uncertainty using
  pomdps,'' in \emph{2019 International Conference on Robotics and Automation
  (ICRA)}.\hskip 1em plus 0.5em minus 0.4em\relax IEEE, 2019, pp. 2751--2757.

\bibitem{lauri2022partially}
M.~Lauri, D.~Hsu, and J.~Pajarinen, ``Partially observable markov decision
  processes in robotics: A survey,'' \emph{IEEE Transactions on Robotics},
  2022.

\bibitem{pajarinen2017robotic}
J.~Pajarinen and V.~Kyrki, ``Robotic manipulation of multiple objects as a
  pomdp,'' \emph{Artificial Intelligence}, vol. 247, pp. 213--228, 2017.

\bibitem{li2016act}
J.~K. Li, D.~Hsu, and W.~S. Lee, ``Act to see and see to act: Pomdp planning
  for objects search in clutter,'' in \emph{2016 IEEE/RSJ International
  Conference on Intelligent Robots and Systems (IROS)}.\hskip 1em plus 0.5em
  minus 0.4em\relax IEEE, 2016, pp. 5701--5707.

\bibitem{xiao2019online}
Y.~Xiao, S.~Katt, A.~ten Pas, S.~Chen, and C.~Amato, ``Online planning for
  target object search in clutter under partial observability,'' in \emph{2019
  International Conference on Robotics and Automation (ICRA)}.\hskip 1em plus
  0.5em minus 0.4em\relax IEEE, 2019, pp. 8241--8247.

\bibitem{invigorate}
H.~Zhang, Y.~Lu, C.~Yu, D.~Hsu, X.~La, and N.~Zheng, ``Invigorate: Interactive
  visual grounding and grasping in clutter,'' \emph{arXiv preprint
  arXiv:2108.11092}, 2021.

\bibitem{yang2022interactive}
Y.~Yang, X.~Lou, and C.~Choi, ``Interactive robotic grasping with
  attribute-guided disambiguation,'' in \emph{2022 International Conference on
  Robotics and Automation (ICRA)}.\hskip 1em plus 0.5em minus 0.4em\relax IEEE,
  2022, pp. 8914--8920.

\bibitem{he2016deep}
K.~He, X.~Zhang, S.~Ren, and J.~Sun, ``Deep residual learning for image
  recognition,'' in \emph{Proceedings of the IEEE conference on computer vision
  and pattern recognition}, 2016, pp. 770--778.

\bibitem{ren2015faster}
S.~Ren, K.~He, R.~Girshick, and J.~Sun, ``Faster r-cnn: Towards real-time
  object detection with region proposal networks,'' \emph{Advances in neural
  information processing systems}, vol.~28, 2015.

\bibitem{girshick2015fast}
R.~Girshick, ``Fast r-cnn,'' in \emph{Proceedings of the IEEE international
  conference on computer vision}, 2015, pp. 1440--1448.

\bibitem{diuk2008object}
C.~Diuk, A.~Cohen, and M.~L. Littman, ``An object-oriented representation for
  efficient reinforcement learning,'' in \emph{Proceedings of the 25th
  international conference on Machine learning}, 2008, pp. 240--247.

\bibitem{roi-grasp}
H.~Zhang, X.~Lan, S.~Bai, X.~Zhou, Z.~Tian, and N.~Zheng, ``Roi-based robotic
  grasp detection for object overlapping scenes,'' in \emph{2019 IEEE/RSJ
  International Conference on Intelligent Robots and Systems (IROS)}.\hskip 1em
  plus 0.5em minus 0.4em\relax IEEE, 2019, pp. 4768--4775.

\bibitem{chang2015shapenet}
A.~X. Chang, T.~Funkhouser, L.~Guibas, P.~Hanrahan, Q.~Huang, Z.~Li,
  S.~Savarese, M.~Savva, S.~Song, H.~Su, \emph{et~al.}, ``Shapenet: An
  information-rich 3d model repository,'' \emph{arXiv preprint
  arXiv:1512.03012}, 2015.

\bibitem{calli2017yale}
B.~Calli, A.~Singh, J.~Bruce, A.~Walsman, K.~Konolige, S.~Srinivasa, P.~Abbeel,
  and A.~M. Dollar, ``Yale-cmu-berkeley dataset for robotic manipulation
  research,'' \emph{The International Journal of Robotics Research}, vol.~36,
  no.~3, pp. 261--268, 2017.

\bibitem{bridson2007fast}
R.~Bridson, ``Fast poisson disk sampling in arbitrary dimensions.''
  \emph{SIGGRAPH sketches}, vol.~10, no.~1, p.~1, 2007.

\end{thebibliography}

\end{document}